\documentclass[twocolumn]{galbot}

\usepackage{wrapfig}
\usepackage{tabularx}
\usepackage{textcomp}
\usepackage{stfloats}
\usepackage{url}
\usepackage{verbatim}
\usepackage{graphicx}
\usepackage{titlesec}
\usepackage{adjustbox}
\usepackage{multirow}
\usepackage{tikz}
\usepackage{comment}
\usepackage{amsmath,amssymb}
\usepackage{colortbl}
\usepackage{color}
\usepackage{booktabs} 
\usepackage{hyperref}
\usepackage{subcaption} 
\usepackage{arydshln}
\RequirePackage{xspace}
\makeatletter
\DeclareRobustCommand\onedot{\futurelet\@let@token\@onedot}
\def\@onedot{\ifx\@let@token.\else.\null\fi\xspace}

\makeatother

\usepackage{makecell}

\usepackage{pifont}
\usepackage{bbding}
\usepackage{fontawesome}
\usepackage{xspace}

\usepackage{float}

\newlength\savewidth

\newcolumntype{x}[1]{>{\centering\arraybackslash}p{#1pt}}
\newcolumntype{y}[1]{>{\raggedright\arraybackslash}p{#1pt}}
\newcolumntype{z}[1]{>{\raggedleft\arraybackslash}p{#1pt}}

\renewcommand{\paragraph}[1]{\vspace{1mm}\noindent\textbf{#1}}

\usepackage{xcolor}
\usepackage{array}
\usepackage{bbm}
\usepackage{collcell,xfp}
\usepackage{pgf}
\usepackage[most]{tcolorbox}
\usepackage{csquotes}
\usepackage[noorphans,vskip=1em,leftmargin=1em]{quoting}
\usepackage{enumitem} 
\usepackage{forest}
\usepackage{caption}
\usepackage{longtable}
\usepackage[T1]{fontenc}

\renewcommand{\paragraph}[1]{\vspace{1.25mm}\noindent\textbf{#1}}

\usepackage{algorithm}
\usepackage{listings}

\definecolor{codeblue}{rgb}{0.25, 0.5, 0.5}
\definecolor{codekw}{rgb}{0.35, 0.35, 0.75}
\lstdefinestyle{Pytorch}{
    language = Python,
    backgroundcolor = \color{white},
    basicstyle = \fontsize{9pt}{8pt}\selectfont\ttfamily\bfseries,
    columns = fullflexible,
    aboveskip=1pt,
    belowskip=1pt,
    breaklines = true,
    captionpos = b,
    commentstyle = \color{codeblue},
    keywordstyle = \color{codekw},
}

\definecolor{green}{HTML}{009000}
\definecolor{red}{HTML}{ea4335}

\definecolor{linecolor1}{RGB}{246, 248, 239}
\definecolor{linecolor2}{RGB}{230, 234, 217}
\definecolor{linecolor3}{RGB}{211, 222, 190}

\usepackage{fix-cm}
\usepackage{orcidlink}
\usepackage{siunitx}

\newcommand{\method}{UniPart}
\newcommand{\objaversegeneral}{Objaverse-General}
\newcommand{\shapenetpart}{ShapeNetPart}

\newcommand{\partnete}{PartNet-E}

\title{UniPart: Towards Zero-shot Language-Grounded 3D Part Segmentation for Embodied Interaction}

\author[1,4,6,*]{Xinqiang Yu\,\orcidlink{0000-0001-8440-8975}}
\author[3,4,*]{Zekun Qi\,\orcidlink{0009-0001-2554-5141}}
\author[6]{Jiawei He\,\orcidlink{0000-0001-6872-3254}}
\author[5]{Wenyao Zhang\,\orcidlink{0009-0006-3090-255X}}
\author[3,4]{Xuchuan Chen\,\orcidlink{0009-0006-7072-4305}}
\author[6]{Guocai Yao\,\orcidlink{0000-0002-0673-0229}}
\author[3,\dagger]{Li Yi\,\orcidlink{0000-0002-9319-0354}}
\author[1,\dagger]{Zhaoxiang Zhang\,\orcidlink{0000-0003-2648-3875}}
\author[2,4,\dagger]{He Wang\,\orcidlink{0000-0002-3365-4620}}

\affiliation[1]{NLPR, Institute of Automation, Chinese Academy of Sciences, China}
\affiliation[2]{Peking University, China}
\affiliation[3]{Tsinghua University, China}
\affiliation[4]{Galbot Inc., China}
\affiliation[5]{Shanghai Jiao Tong University, China}
\affiliation[6]{Beijing Academy of Artificial Intelligence, China}
\contribution[*]{Equal contribution}
\contribution[\dagger]{Corresponding authors}
\galbotdata[Email]{{\scriptsize\texttt{\{yuxinqiang,qizekun,chenxuchuan,yili,wanghe\}@galbot.com}}\\
\texttt{\{jwhe2024,wyzhang2022\}@gmail.com}\\
\texttt{gcyao1@baai.ac.cn}; \texttt{zhaoxiang.zhang@ia.ac.cn}}

\hypersetup{
  pdftitle={UniPart: Towards Zero-shot Language-Grounded 3D Part Segmentation for Embodied Interaction},
  pdfauthor={Xinqiang Yu, Zekun Qi, Jiawei He, Wenyao Zhang, Xuchuan Chen, Guocai Yao, Li Yi, Zhaoxiang Zhang, He Wang},
  pdfkeywords={Zero-Shot, 3D Part Segmentation, Embodied Interaction}
}

\page{\url{https://xinqiangyu.github.io/UniPart/}}

\abstract{
Fine-grained robotic manipulation depends on understanding parts, not only whole objects. Existing 3D foundation models tend to be either generalized but object-aware, or part-aware but limited to closed-set taxonomies, which weakens zero-shot transfer. We study text-conditioned 3D part segmentation, where a free-form phrase selects a functional part on point cloud. We introduce UniPart, a feed-forward cross-modal 3D Transformer that conditions CLIP text embedding. 
To scale supervision, we build LangPart-1M with 160K+ Objaverse assets and 8M text to part pairs using multi-view consistent part generation. 
We further manually label a high-quality subset, LangPart-4K, for fine-tuning and evaluation.
UniPart achieves strong zero-shot results on open-vocabulary part benchmarks and transfers to language-conditioned part grasping in real world. 
  \par\smallskip\textbf{Keywords:} Zero-Shot; 3D Part Segmentation; Embodied Interaction
}

\begin{document}
\maketitle
\pagestyle{empty}
\raggedbottom

\begin{figure*}[t]
\centering
\includegraphics[width=0.87\linewidth]{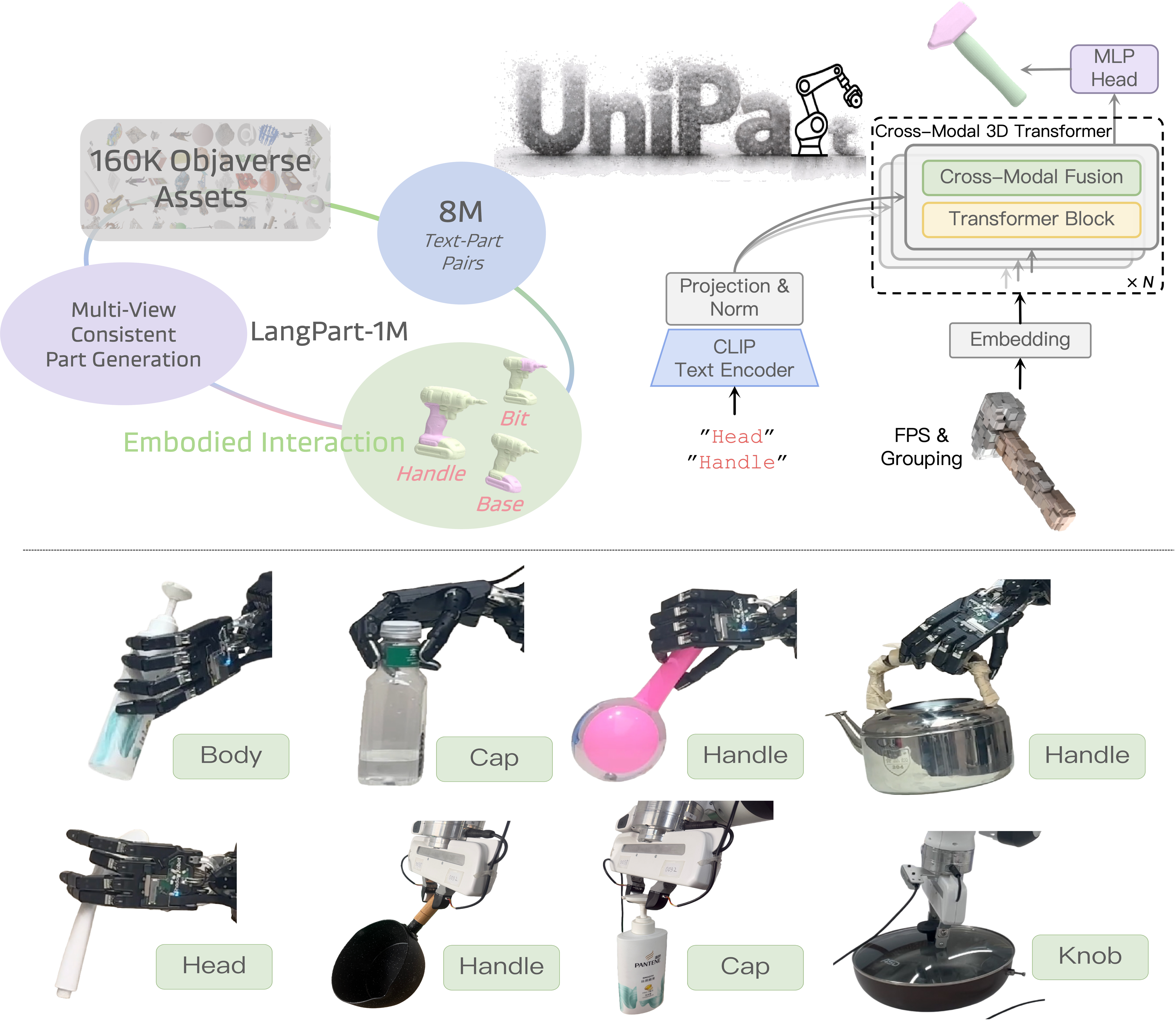}
\caption{\textbf{Overview.} We construct LangPart-1M from 160K+ Objaverse assets and generate 8M text to part pairs using a scalable multi-view consistent part generation pipeline. UniPart is a feed-forward text-conditioned 3D part segmentation model trained on this dataset. The predicted part masks provide fine-grained geometric cues that support embodied interaction.}
\label{fig:teaser}
\end{figure*}

\section{Introduction}
\label{sec:intro}

The quest for truly general-purpose embodied intelligence hinges on a robot's ability to not merely perceive objects, but to comprehend their function and composition at a granular level~\cite{Affordance77,RTAffordance24,ShapeLLM24,affordancellm24}. 
In open-world settings, manipulation depends on generalizing to unseen objects and to the parts that matter for an action. 
A robot must identify the semantically relevant object part on a previously unseen instance, conditioned on a human instruction.

Foundation models such as SAM~\cite{kirillov2023segment} and CLIP~\cite{radford2021learning} have made 2D perception generalizable, and several lines of work extend these priors to 3D. Yet fine-grained part understanding remains difficult~\cite{affordancellm24,3d-affordancellm25,ShapeLLM24}. Existing approaches often fall into two patterns that leave a gap for dexterous manipulation.

\begin{itemize}[leftmargin=1.5em]
\item \textbf{Promptable but coarse} Models that inherit SAM-style prompting in 3D, such as Point-SAM~\cite{zhou2024point}, are strong at separating object instances but do not resolve internal semantically relevant parts within one object.
\item \textbf{Part-aware but closed-set} Part and affordance models trained on datasets such as ShapeNetPart~\cite{ShapeNetPart16} provide fine granularity, yet they rely on fixed taxonomies and struggle with open-vocabulary generalization.
\end{itemize}

We frame the missing capability as a semantic interface for fine-grained 3D geometry. Instead of predicting from a fixed label set, a robot should take a natural language instruction and return the corresponding 3D part mask. We study \emph{language-grounded zero-shot 3D part segmentation} and tackle it with a paired design of supervision and model.

On the supervision side, we build LangPart-1M. The key challenge is not only scale, but also multi-view consistency of part masks when lifting from 2D to 3D. Our pipeline combines multi-view rendering, SAM region proposals, Set-of-Mark prompting, and a text-guided merging step followed by lifted to 3D. This produces point-level part masks aligned to language at scale. We then curate LangPart-4K with manual verification so that training can benefit from both breadth and high-fidelity corrections.
On the model side, we introduce UniPart, which is a feed-forward cross-modal 3D Transformer that injects a frozen CLIP text embedding into every layer through simple addition. This choice keeps the architecture lightweight while maintaining strong language conditioning. 
We also include a controlled study that separates the effect of the dataset from the effect of the model, which is critical for positioning against closely related work.

In summary, our key contributions are:

\begin{itemize}[leftmargin=1.5em]
\item \textbf{LangPart-1M and LangPart-1K} We introduce a large-scale dataset with 160K objects and 8M text-to-part pairs, and a benchmark split with 1K objects and more than 4K annotations for open-vocabulary evaluation.
\item \textbf{Scalable data engine and high-quality subset} We present a multi-view pipeline that aligns language to 3D part masks using SAM proposals, Set-of-Mark prompting, and text-guided merging for view consistency. We additionally build LangPart-4K with ten annotators and assign text labels to part masks, providing high-fidelity language-grounded 3D part supervision.
\item \textbf{UniPart model and controlled positioning} We propose a feed-forward cross-modal 3D Transformer with layer-wise additive text injection, and we report a matched supervision study that disentangles the gains from data and from architecture.
\item \textbf{Embodied validation} We show that UniPart can produce part masks on real sensor point clouds that are useful for language-conditioned grasping.
\end{itemize}

\section{Related Works}
\label{sec:related_works}

\subsection{3D Part Segmentation}
Classic 3D part segmentation relies on closed-set supervision, such as ShapeNetPart~\cite{ShapeNetPart16} and PartNet~\cite{partnet19}, and achieves fine granularity within fixed taxonomies. More recent work leverages 2D vision language priors to reduce labeling cost, yet open-vocabulary generalization is still limited~\cite{PartSLIP22}. In parallel, promptable and open-vocabulary 3D segmentation extends SAM~\cite{SAM23} and CLIP~\cite{CLIP21} into 3D through multi-view lifting or 3D scene pipelines, and it is mainly evaluated on instance or scene semantics rather than object internal parts~\cite{LangSplat24,PointSAM25,COPS25}. Recent efforts also explore open-vocabulary part segmentation in narrower domains or synthetic settings~\cite{OpenPart3D25}.

Our work is closest to concurrent text-prompted part segmentation such as FIND3D~\cite{ma2025find}. We share the goal of querying arbitrary parts with language, while differing in two aspects that we evaluate under matched supervision. First, we construct LangPart-1M using a multi-view pipeline that enforces 3D consistency through text guided mask merging. Second, we train a feed-forward point-based model with layer-wise additive text injection, which provides dense part masks without multi-view querying at inference. We include a controlled study that isolates the effects of supervision and architecture.

Related but distinct, language-guided 3D affordance segmentation focuses on action regions rather than semantic parts and often depends on task-specific definitions~\cite{TAFS24,GEAL24}. UniPart targets category-agnostic part masks as a perception primitive that can be composed into downstream manipulation.

\subsection{Language-Grounded Robot Manipulation}
Language-grounded robot Manipulation adopts the human language as a general instruction interface.
Existing works can be categorized into two groups:
\textbf{(i)} \textit{End-to-end} models like RT-series~\cite{RT123,RT223,RTH24} built upon unified cross-modal Transformers with tokenized actions~\cite{PERACT22,InstructRL22,ALOHA23}, large vision-language-action models built from VLMs~\cite{OpenVLA24,dreamvla25}, or 3D representations~\cite{3DVLA24,RoboPoint24}.
Training on robot data such as Open X-Embodiment~\cite{OpenXEmbodiment24} and DROID~\cite{DROID24}, a remarkable process has been made.
However, the data \textit{scale} is still limited compared to in-the-wild data for training VLMs.
\textbf{(ii)} \textit{Decoupled} high-level reasoning and low-level actions in large VLMs and small off-the-shelf policy models, primitives~\cite{SayCan22,CodeAsPolicy23,VoxPoser23,CoPa24,MOKA24,UniSim24,ManipAnywhere24,robomatrix24,dexvlg25,code_as_monitor25}, or articulated priors~\cite{a3vlm24,ManipLLM24}.
Our model lies in this group, where an open-world generalization property emerges from VLMs and our proposed UniPart is empowered by fine-grained understanding.
\subsection{3D Representation Learning}
Research on 3D Representation Learning encompasses various methods, including point-based~\cite{PointNet17,PointNet++17}, voxel-based~\cite{voxelnet15}, and multiview-based approaches~\cite{MVCNN3D15,MVTN21}. 
Point-based methods~\cite{PointNext22,PointTrans21} have gained prominence in object classification~\cite{ModelNet15,ScanObjectNN19} due to their sparsity yet geometry-informative representation. On the other hand, voxel-based methods~\cite{voxelrcnn21,SyncSpecCNN17,VPP23} offer dense representation and translation invariance, leading to a remarkable performance in object detection~\cite{ScanNet17} and segmentation~\cite{ShapeNetPart16,S3DIS16}.
The evolution of attention mechanisms~\cite{Transformer17} has also contributed to the development of effective representations for downstream tasks, as exemplified by the emergence of 3D Transformers~\cite{PointTrans21,groupfree21,voxeltransformer21}. Notably, 3D self-supervised representation learning has garnered significant attention in recent studies. PointContrast~\cite{PointContrast20} utilizes contrastive learning across different views to acquire discriminative 3D scene representations. Innovations such as Point-BERT~\cite{PointBERT22} and Point-MAE~\cite{PointMAE22} introduce masked modeling~\cite{MAE22,BERT21} pretraining into the 3D domain. 
ACT~\cite{ACT23} pioneers cross-modal geometry understanding through 2D or language foundation models such as CLIP~\cite{CLIP21} or BERT~\cite{BERT21}. 
Following ACT, {\scshape ReCon}~\cite{ReCon23} further proposes a learning paradigm that unifies generative and contrastive learning. 
PPT~\cite{ppt24} highlights the significance of positional encoding in 3D representation learning.
Additionally, leveraging foundation vision-language models like CLIP~\cite{ACT23,CLIP21} has spurred the exploration of a new direction in open-world 3D representation learning. This line of work seeks to extend the applicability and adaptability of 3D representations in diverse and open-world/vocabulary scenarios~\cite{OpenScene23,CLIPFO3D23,Lowis3D23,PLA23,PointGCC23,Zhang_2025_ICCV}.

\section{UniPart: Language-Grounded 3D Part Segmentation for Embodied Interaction}
\label{sec:models}

\begin{table*}[t!]
\centering
\caption{Properties and inference time of \method~and baselines. 
We report whether each method is open-world, supports cross-category transfer, predicts part-level masks, and runs as a feed-forward 3D model without multi-view rendering and backprojection at inference. We also report a representative runtime and the number of training objects when applicable.
}

\label{table:method_properties}
\small
\begin{adjustbox}{max width=\linewidth}
\begin{tabular}{l|c|c|c|c|c|c}
\toprule
Method & Time & Open-world & Cross-category & Part-level & Feed-forward & Num. Object\\
\midrule
PointCLIPV2 & 5.4s & \ding{51} & \ding{51} & \ding{51} & \ding{55} & - \\
PartSLIP++ & 174.3s & \ding{51} & \ding{55} & \ding{51} & \ding{55} & -\\
OpenMask3D & 296.5s & \ding{51} & \ding{51} & \ding{55} & \ding{55} & -\\
PartDistill & 0.7s & \ding{55} & \ding{55} & \ding{51} & \ding{51} & -\\
PointNext  & 1.4s & \ding{55} & \ding{51} & \ding{51} & \ding{51} & -\\
FIND3D & 0.9s & \ding{51} & \ding{51} & \ding{51} & \ding{51} & 36k\\
\method(Ours) & \textbf{0.4s} & \ding{51} & \ding{51} & \ding{51} & \ding{51} & \textbf{160k+} \\
\bottomrule
\end{tabular}
\end{adjustbox}

\end{table*}

\subsection{Datasets}
Scaling text-part alignment dataset to hundreds of thousands of 3D assets requires balancing annotation scalability and quality. Fully manual labeling at this scale is impractical, whereas fully automated annotation inevitably introduces noise.
To address this trade-off, we construct our dataset in two complementary stages. The first subset \textbf{LangPart-1M} is generated through a fully automated pipeline leveraging large-scale foundation models, providing broad coverage and diversity across object categories. While scalable, this subset may contain minor inconsistencies due to model-induced noise.
To provide high-fidelity semantic supervision, we additionally curate a manually annotated subset \textbf{LangPart-4K} with carefully verified part labels, which is splited into \textbf{LangPart-3K} for training and \textbf{LangPart-1K} as benchmark.

During training, we adopt a \textbf{three-stage curriculum}. We first align point-cloud representations with RGB features to inject rich cross-modal semantic cues into the 3D backbone, establishing an initialization that is both geometrically grounded and semantically aware. 
We then pretrain the model on large-scale data to acquire broad structural and semantic priors at scale. Finally, we fine-tune on the manually annotated subset to correct the noise of automatic supervision and sharpen part-level localization accuracy. This staged design progressively transfers knowledge from cross-modal semantic alignment to scalable representation learning, and ultimately to high-fidelity part understanding.

\subsubsection{Dataset Statistics}\label{sec:dataset-stats}
To facilitate rigorous evaluation and training, LangPart-1M is curated with extensive coverage and fine-grained annotations. Specifically, the dataset contains over 160K+ unique 3D objects spanning more than 250 distinct object categories (e.g., furniture, tools, appliances, vehicles), as shown in \cref{table:method_properties}. Within these objects, we annotate over 1,000 unique part types (e.g., handle, button, spout, screen), yielding a total of approximately 8 million text-part pairs.
We train on the remaining data after holding out the benchmark split, and report results on our proposed benchmark.

\subsubsection{Data Source}\label{sec:objaverse-so}

To enable large-scale training, we construct the LangPart-1M dataset based on Objaverse~\cite{objaverse23}, which contains approximately 800K Internet-sourced 3D models across diverse categories.
However, Internet-sourced models often exhibit noisy annotations and inconsistent visual quality. To obtain clean and consistent supervision, we re-render the assets in Blender with controlled lighting, generating over 8M high-fidelity images.

\subsubsection{Data Filtering} \label{sec:data-filtering}
To clean the data that can be better used for generating fine-grained part annotations, we first clean the data by using a dedicated filtering strategy that filters data to preserve samples that satisfy the following six requirements.
\ding{182} Clean objects without the ground for auxiliary visualization.
\ding{183} Reasonable objects that have sufficient spatial reasoning potentials.
\ding{184} High-quality objects. Low-quality objects such as blurry and wrong samples are filtered.
\ding{185} Distinguishable objects. Abstract objects such as meaningless solids are filtered.
\ding{186} Non-scene objects. Samples that describe a 3D scene are filtered for object-centric understanding purposes.
\ding{187} No test leakage. We proactively remove all asset IDs from evaluation benchmarks, including Objaverse-General, from the training pool.

Filtering at this scale is impractical to do fully by hand. Inspired by recent work on using VLMs as image-based judges~\cite{LLMAsJudge23,GPT4V-3DEvaluator24,DreamBenchPlus24}, we use GPT-5.2-Pro~\cite{gpt5_26} with a fixed prompt and deterministic decoding to decide whether an asset passes the criteria. For reproducibility, we will release the full filtering prompt, the list of retained asset IDs, and the per-asset decisions so that the training set can be reconstructed without re-running the VLM.

\subsubsection{LangPart-1M Dataset Generation}
\label{sec:annotation-pipeline}
\begin{figure*}[t!]
  \begin{center}
  \includegraphics[width=1.0\linewidth]{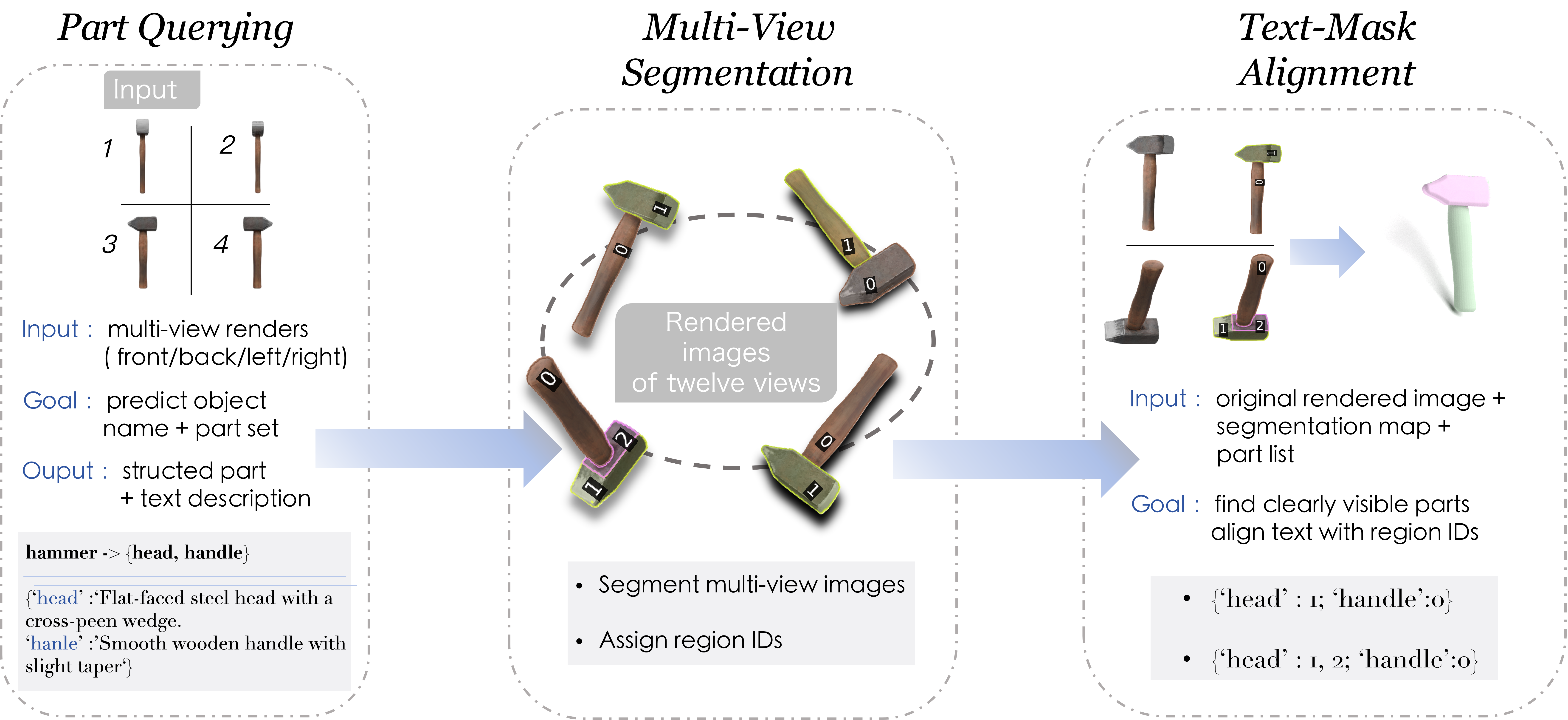}
  \caption{Pipeline of \textbf{LangPart-1M} dataset generation, including Part Querying, Multi-View Segmentation and Text-Mask Alignment.
  } \label{fig:data_pipeline}
  
  \end{center}
\end{figure*}
To generate the fine-grained annotations for our dataset, we devise a novel pipeline that systematically leverages large-scale foundation models. As shown in \cref{fig:data_pipeline}, this process follows four stages:

\vspace{3pt}
\noindent(i)~\textbf{Part-List Querying:} For each cleaned 3D object, we first employ GPT-5.2-Pro~\cite{gpt5_26} to query and generate its object name and a comprehensive list of its potential functional or meaningful parts.

\vspace{3pt}
\noindent(ii)~\textbf{2D Unsupervised Segmentation}
For each object, we render twelve randomly sampled single-view images in Blender. For each view, we apply SAM~\cite{SAM23} to obtain all possible segmentations, producing multiple fine-grained part candidates per image. To enable large language models to reason about these spatial regions, we employ the Set-of-Mark~\cite{SoM23} prompting technique, assigning a unique visual identifier to each generated segment.

\vspace{3pt}
\noindent(iii)~\textbf{Multi-View Consistent Part Generation} \label{sec:part-generate}
Lifting 2D masks into 3D often produces view-dependent inconsistencies. We address this with a text guided merging step followed by lifted to 3D. For each view, we provide the SoM labeled image to GPT-5.2-Pro and ask it to select the visible parts from the queried part list and to merge the region IDs that belong to the same part. 
This step improves cross-view coherence, merges over-segmented regions, and suppresses isolated outliers.

\vspace{3pt}
\noindent(iv)~\textbf{Text to Point Cloud Alignment} Finally, using the camera intrinsics, extrinsics, and depth maps from our renderer, we aggregate the back-projected points and obtain a unified point cloud with per-point part labels. To support reproducibility, we will release the intermediate artifacts needed to replay the pipeline, including render metadata, depth maps, SAM masks, SoM overlays, and the merged region sets produced during alignment.

\subsubsection{Manual Annotation for Quality Improvement}\label{qua_imp}
After data filtering in Sec.~\ref{sec:data-filtering}, we obtain over 160K 3D assets from Objaverse. However, not all objects are equally relevant for functional robotic manipulation. We therefore manually select 4K objects and manually construct and label a high-quality subset for post-training, named \textbf{LangPart-4K} by two stages.

\vspace{3pt}
\noindent(i)\textbf{Part-Aware Object Filtering}
In practice, robots primarily interact with objects that exhibit clear functional or structurally meaningful parts. Therefore, from the 160K objects, we further select 4K objects that contain \emph{explicit functional components or well-defined part structures}. The subset is manually annotated to provide high-fidelity supervision for the third training stage.

\vspace{3pt}
\noindent(ii)\textbf{Refinement of Auto Annotation}
For each object in LangPart-4K, we start from the automatically generated multi-view masks from Sec.~\ref{sec:part-generate}. Annotators inspect the correspondence between each part name and its regions in every rendered view and correct mismatches when needed. This yields a verified set of part masks that is used for fine-tuning and for constructing our benchmark.

\begin{figure*}[t!]
  \begin{center}
  \includegraphics[width=0.97\linewidth]{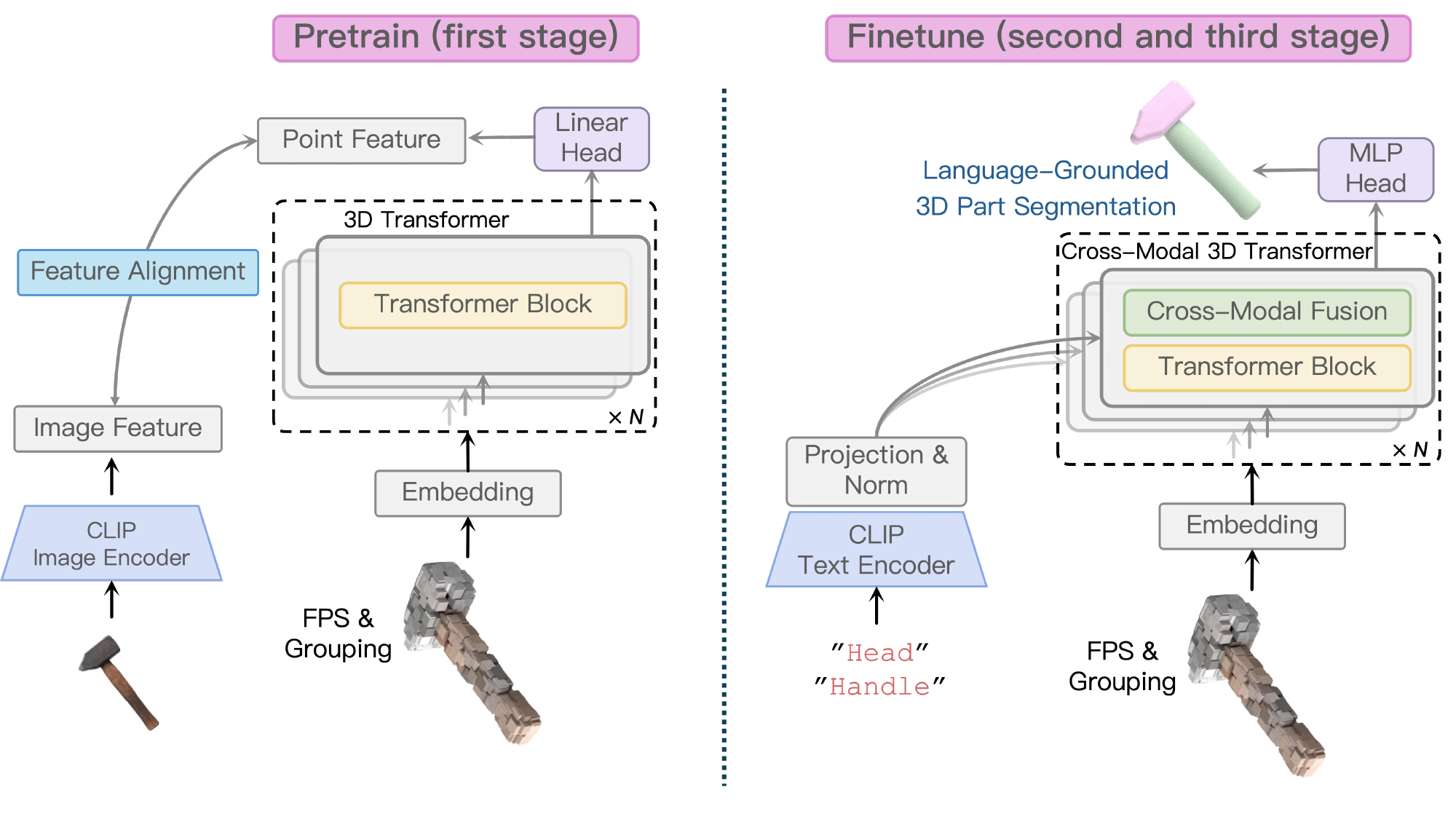}
  \caption{\textbf{UniPart architecture} for part segmentation. During the pretraining stage, UniPart is trained to align image feature and point feature. And for the finetuning stage, UniPart takes 3D \& text embeddings as inputs of a $N$-layer plain Transformer model, and uses a token-level sum as cross-modal fusion in all layers. A PointNet decoder predicts the logits per point.
  } \label{fig:UniPart}
  
  \end{center}
\end{figure*}

\subsubsection{Text-Alignment Part Segmentation Benchmark}
Existing 3D part segmentation benchmarks are limited in diversity and flexibility. They cover only a small set of object categories, rely on closed-set part taxonomies, and often contain canonicalized CAD models. ShapeNetPart has 16 categories and 41 part labels~\cite{ShapeNetPart16}. PartNet-E expands to 45 categories but still uses a fixed vocabulary shared across categories~\cite{partnet19}.
After manual verification in Sec.~\ref{qua_imp}, we split LangPart-4K into LangPart-3K for training and LangPart-1K with 1K objects for evaluation. LangPart-1K is designed to evaluate both seen and unseen categories and part types, and we will release the split specification together with the asset IDs and text prompts used for evaluation.

\subsection{A Cross-Modal 3D Transformer for Language-Grounded Part Segmentation}
We introduce UniPart, a simple Transformer-based architecture~\cite{Transformer17} with cross-modal point cloud–language fusion for open-vocabulary text-conditioned part segmentation. As illustrated in \cref{fig:UniPart}, UniPart takes an object’s point cloud and a language description as input and predicts the corresponding text-conditioned segmentation mask.

\subsubsection{Language and Point Cloud Embeddings}
Given an object point cloud $X = \{\mathbf{x}_i \in\mathbb{R}^3|i=1,2,\dots, N\}$ in Cartesian coordinates and an arbitrary language description $\ell$, we first tokenize both modalities into discrete embeddings.
For the 3D input, following~\cite{ACT23,PointBERT22,ReCon23}, we sample 
$N_s$ seed points via Farthest Point Sampling (FPS) and form local neighborhoods using $k$-nearest Neighbors (kNN). Each neighborhood is encoded into a local geometric token using a lightweight PointNet-style feature extractor~\cite{PointNet17,PointNet++17}, yielding a sequence of 3D tokens.
For language, we adopt OpenAI CLIP~\cite{CLIP21} and use its global text embedding as the conditioning signal for cross-modal fusion with the 3D tokens. The fused representations are then processed by a Transformer encoder.
Finally, a MLP-style decoder head upsamples {\color{gray}[CLS]} token features back to per-point predictions and outputs text-conditioned binary mask over the input points, producing an open-vocabulary part segmentation mask corresponding to $\ell$.
Similar to Vision Transformer~\cite{ViT21}, we adopt a Large configuration with more Transformer layers, larger hidden dimensions, and a stronger CLIP text backbone.

\subsubsection{Cross-Modal Fusion}
We adopt a dense, layer-wise fusion strategy that injects the global text embedding into each layer of the 3D Transformer. In principle, cross-modal fusion can be implemented in many ways, e.g., cross-attention, adapters, or feature concatenation along the channel dimension. Empirically, we find that a simple additive injection—adding the text feature to every point token at each layer—yields the best trade-off between simplicity and effectiveness. We hypothesize this is because the language input is relatively short, and per-token addition strengthens language conditioning throughout the network.

\subsection{Optimization}
Let $\mathcal{F}_{\text{UP}}$ represent the \method~model parameterized by $\theta_{\text{UP}}$ (the CLIP is kept frozen and thus its parameters are not included).
Given every object point cloud $X_i \in \mathcal{D}_{\text{LangPart-1M}}$ in the LangPart-1M dataset, where each object is labeled with a part text set $L_i=\{\ell_j^i, j=1,2,\dots,P_{i}\}$
where $P_{i}$ is the part number of point cloud $X_i$.
And the corresponding ground truth 3D part segmentation mask $S_i=\{\mathbf{s}^i_j, j=1,2,\dots,P_{i}\}$
where $S_{i}$ is an one-hot label. For the second and third \textbf{finetune stage}, the optimization is to minimize the Binary Cross-Entropy (BCE) loss between predicted segmentation mask and ground truth label, where $\mathcal{L}_{\text{BCE}}(\hat{S}_j^i, S_j^i) = - [S_j^i \log(\hat{S}_j^i) + (1 - S_j^i) \log(1 - \hat{S}_j^i)]$.
And the final optimization function is 
\begin{equation}
\scalebox{0.95}{$
     \min_{\theta_{\text{UP}}} \sum_{X_i \in \mathcal{D}_{\text{LangPart-1M}}} \sum_{S_j^i\in S_i} \mathcal{L}_{\text{BCE}}(\hat{S}_j^i, S_j^i)
$}
\end{equation}
During the first \textbf{pretraining stage}, for each single view point cloud  $X^{sin}_i \in \mathcal{D}_{\text{LangPart-1M}}$, and a corresponding RGB image $I_i$ rendered from the same view, and a mapping $\pi(i) \in \{ 1,...,M\}$ that assigns each point to an image patch index, where $M$ is the image patch number. We extract patch-level CLIP features from the image $ \{ c_j\}_{j=1}^{M}$and align them with per-point features produced by the point encoder, encouraging the 3D representation to inherit CLIP’s semantic structure. This initialization significantly facilitates subsequent text-guided part understanding during fine-tuning. And the loss function is 
\begin{equation}
\scalebox{0.95}{$
    \mathcal{L}_{\text{align}}
= \max\!\left(0,\; 1 - \frac{1}{N}\sum_{i=1}^{N}
\left\langle \hat{\mathbf{z}}_{i},\; \hat{\mathbf{c}}_{\pi(i)} \right\rangle \right)
$}
\end{equation}

where $\hat{\mathbf{z}}_{i}$ is the normalized point feature, $\hat{\mathbf{c}}_{\pi(i)}$ is its corresponding normalized CLIP patch feature, $N$ is the point number, and $\langle \cdot, \cdot \rangle$ denotes the inner product.

\subsection{Comparison with Related Works}
Recent work such as FIND3D~\cite{ma2025find} also studies text-prompted 3D part segmentation. We focus on two differentiators that we evaluate under matched supervision. LangPart-1M provides large-scale point-level supervision constructed with a multi-view consistency mechanism based on text-guided merging. We report a controlled study that isolates the contribution of the supervision engine and the contribution of the architecture, and we test the resulting part masks in a real robotic grasping pipeline.

\section{Experiments}
\label{sec:experiments}

\subsection{Experimental Settings}

\vspace{3pt}
\noindent\textbf{Benchmarks}
We evaluate on three settings. We report results on \textbf{Objaverse-General}~\cite{ma2025find} to compare with prior open-world baselines. We additionally evaluate on \textbf{ShapeNetPart}~\cite{ShapeNetPart16} and PartNet-E~\cite{partnet19} to test transfer to classic part benchmarks. Finally, we evaluate on \textbf{LangPart-1K}, which is constructed from LangPart-4K with manual verification.

\vspace{3pt}
\noindent\textbf{Metric}
We report mean Intersection over Union. For each input point cloud, we compute the mean IoU across its annotated parts, then average over the point cloud.

\subsection{Part Segmentation Baselines}
\subsubsection{Open-world Baselines}
To ensure a consistent and fair comparison, we follow the baseline selection and evaluation protocol in FIND3D~\cite{ma2025find} and detail the adopted methods below. We use the same two prompt formats, a compositional query of the form \texttt{\{part\} of a \{object\}} and a short query \texttt{\{part\}}.

\vspace{3pt}
\noindent \textbullet~\textbf{PointCLIPV2}~\cite{PointCLIPV2_23} is an open-world 2D-to-3D framework that repeatedly queries CLIP~\cite{CLIP21} across rendered views. It relies on a prompt-selection procedure that retains the top-$k$ textual prompts for each object, with $k = 1400 \times n_{\text{parts}}$, where the prompt set is chosen using the ShapeNetPart test split. 

\vspace{3pt}
\noindent \textbullet~\textbf{PartSLIP++}~\cite{liu2023partslip} follows a detection-driven design, leveraging GLIP~\cite{li2022grounded} together with a dedicated superpoint generation module to transfer 2D detections back to 3D. The method additionally trains category-specific models on PartNet-E~\cite{mo2019partnet}; for consistency and fairness, we report results using its released zero-shot checkpoint. 

\vspace{3pt}
\noindent \textbullet~\textbf{OpenMask3D}~\cite{takmaz2023openmask3d} is an open-vocabulary 2D-to-3D approach originally trained on 3D scenes. Both PointCLIPV2 and OpenMask3D produce dense predictions by assigning a label to every point. When the benchmark includes unlabeled regions, we add an ``other'' text query to represent the no-label option.

\begin{table*}[t!]
\caption{Performance comparison with open-world methods on \objaversegeneral. UniPart performs best on Objaverse-General benchmark.
We show results evaluated with 2 common query prompts: ``\{part\} of a \{object\}" and ``\{part\}" for all methods following FIND3D~\cite{ma2025find}.}

\label{table:miou_compare_objaverse}
\begin{center}
\setlength{\tabcolsep}{6pt}
\small
\begin{adjustbox}{max width=\linewidth}
\begin{tabular}{l|cc|cc}
\toprule
Methods & \multicolumn{4}{c}{\objaversegeneral} \\
\midrule
 & \multicolumn{2}{c|}{Seen Categories} & \multicolumn{2}{c}{Unseen Categories} \\
\midrule
  & {\{part\} of a \{object\}}  & \{part\}  & {\{part\} of a \{object\}} & \{part\} \\
\midrule
PointCLIPV2 & 9.81 & 11.27 & 10.27 & 11.09 \\
PartSLIP++ & 2.69 & 15.03 & 0.57 & 10.43 \\
OpenMask3D & 11.81 & 11.93 & 7.01 & 10.31 \\
FIND3D & 33.78 & 34.10 & 26.21 & 27.41 \\
\textbf{\method (ours)} &  \textbf{46.43} & \textbf{49.27} & \textbf{41.46} & \textbf{45.23} \\
\bottomrule
\end{tabular}
\end{adjustbox}

\end{center}
\end{table*}
\vspace{3pt}

\begin{table*}[t!]
\caption{Performance comparison with open-world methods on \shapenetpart. UniPart also performs best on this benchmark. And we also evaluate with prompts ``\{part\} of a \{object\}" and ``\{part\}".}

\label{table:miou_compare_shapenet}
\begin{center}
\setlength{\tabcolsep}{4pt}
\small
\begin{adjustbox}{max width=\linewidth}
\begin{tabular}{l|cc|cc|cc}
\toprule
Methods & \multicolumn{6}{c}{\shapenetpart} \\
\midrule
 & \multicolumn{2}{c|}{Canonical} & \multicolumn{2}{c|}{Rotated} & \multicolumn{2}{c}{ShapeNetPart-V2} \\
\midrule
  & {\{part\} of a \{object\}} & \{part\}  & {\{part\} of a \{object\}} & \{part\} & {\{part\} of a \{object\}} & \{part\} \\
\midrule
PointCLIPV2 & 16.91 & 20.22 & 16.88 & 18.19 & 15.14 & 17.11 \\
PartSLIP++ & 1.43 & 6.46 & 0.94& 6.03 & 1.54 & 11.62 \\
OpenMask3D & 8.94 & 10.37 & 6.75 & 14.56 & 15.87 & 13.77\\
FIND3D & 28.39 & 24.09 & 29.64 & 23.71 & 42.15 & 30.02\\
\textbf{\method (ours)} & \textbf{40.07} & \textbf{41.75} & \textbf{42.43} & \textbf{39.56} & \textbf{55.63} & \textbf{51.29}\\
\bottomrule
\end{tabular}
\end{adjustbox}
\vspace{3pt}

\end{center}
\end{table*}

\subsection{Experimental Results}
\begin{table*}[t!]
\caption{Comparison on LangPart-1K benchmark.}

\label{table:langpart-bench-exp}
\begin{center}
\setlength{\tabcolsep}{6pt}
\small
\begin{adjustbox}{max width=\linewidth}
\begin{tabular}{l|SS|SS}
\toprule
Methods & \multicolumn{4}{c}{LangPart-1K benchmark} \\
\midrule
& \multicolumn{2}{c|}{Single-View point cloud} & \multicolumn{2}{c}{Whole Object point cloud} \\
\midrule
& {\{part\} of a \{object\}} & \{part\} &
   {\{part\} of a \{object\}} & \{part\} \\
\midrule
FIND3D        & 10.39 & 8.97 & 14.36 & 17.28 \\
\textbf{\method(ours)} & \textbf{31.23} & \textbf{33.56} & \textbf{28.69} & \textbf{27.17} \\
\bottomrule
\end{tabular}
\end{adjustbox}
\vspace{1pt}

\end{center}
\end{table*}

\vspace{3pt}
\noindent\textbf{Results on Objaverse-General, ShapeNetPart, and PartNet-E}
Table~\ref{table:miou_compare_objaverse} and Table~\ref{table:miou_compare_shapenet} report mIoU on Objaverse-General and ShapeNetPart. As shown in this table, our method UniPart can perform better than others.
As shown in Table~\ref{table:miou_partnete}, we compare open-world part segmentation methods on \partnete. 
Among these approaches, Our method has best performance.

\begin{figure*}[t]
\centering
\includegraphics[width=\linewidth]{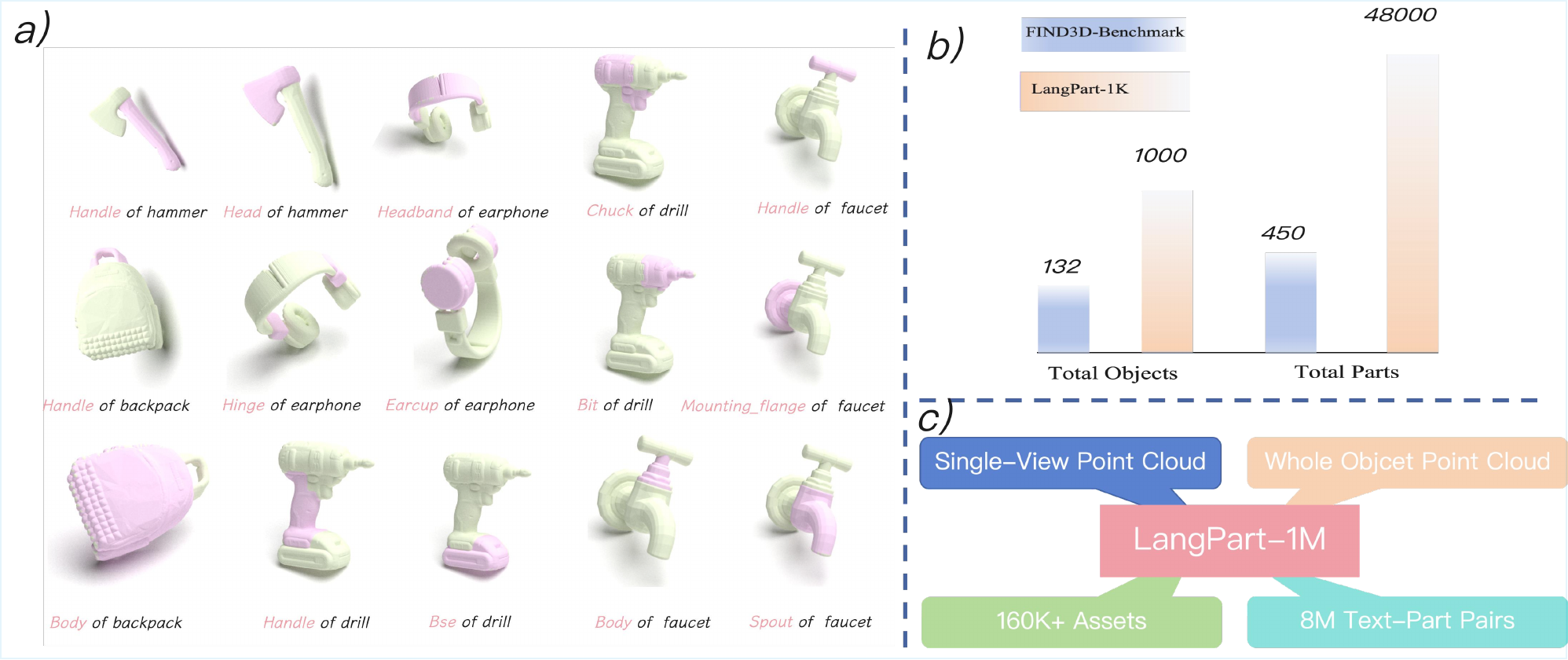}
\captionof{figure}{\textbf{Qualitative results and our dataset.} a) Examples of our dataset. b) Our benchmark has more objects and parts than FIND3D. c) Our datasets has single-view or whole point cloud, more objects and more pairs.}
\label{fig:big_vis}

\end{figure*}

\vspace{3pt}
\noindent\textbf{Qualitative results}
Figure~\ref{fig:big_vis} shows qualitative examples from our model predictions. Our benchmark and dataset has more objects and more parts than FIND3D. And our datasets include single-view and whole point cloud.

\vspace{3pt}
\noindent\textbf{Results on LangPart-1K}
\begin{table*}[t]
\centering
\caption{Comparison of open-world methods on \partnete. 
}

\label{table:miou_partnete}
\begin{center}
\setlength{\tabcolsep}{5pt}
\small
\begin{adjustbox}{max width=\linewidth}
\begin{tabular}{l|c|c|c|c}
\toprule
Methods & \multicolumn{2}{c|}{Canonical Orientation} & \multicolumn{2}{c}{Rotated}\\
\midrule
& \{part\} of a \{object\} & \{part\} & \{part\} of a \{object\} & \{part\} \\
\midrule
Find3D & 16.86 & 16.38 & 17.62 & 17.16 \\
PartSLIP++ & \phantom{0}5.12 & 32.71 & \phantom{0}3.87 & 23.03 \\
PointCLIPV2 & 11.28& \phantom{0}9.70 & 10.32 & 10.22 \\
OpenMask3D & 12.54 & 11.24 & 11.93 & 11.67 \\
\textbf{\method (ours)} & \textbf{23.15} & \textbf{22.90} & \textbf{24.28} & \textbf{21.72} \\
\bottomrule
\end{tabular}
\end{adjustbox} 
\end{center}

\end{table*}
Our dataset provides both single-view point clouds and complete 3D point clouds. Accordingly, we train two separate models—one tailored to partial observations and the other operating on full geometry. For a fair and comprehensive evaluation, we further report results under both test protocols, as shown in Table~\ref{table:langpart-bench-exp}. And Our method has best performance.

\subsection{Fine-Grained Robotic Manipulation}

\begin{figure*}[t]
\centering
\includegraphics[width=0.88\linewidth]{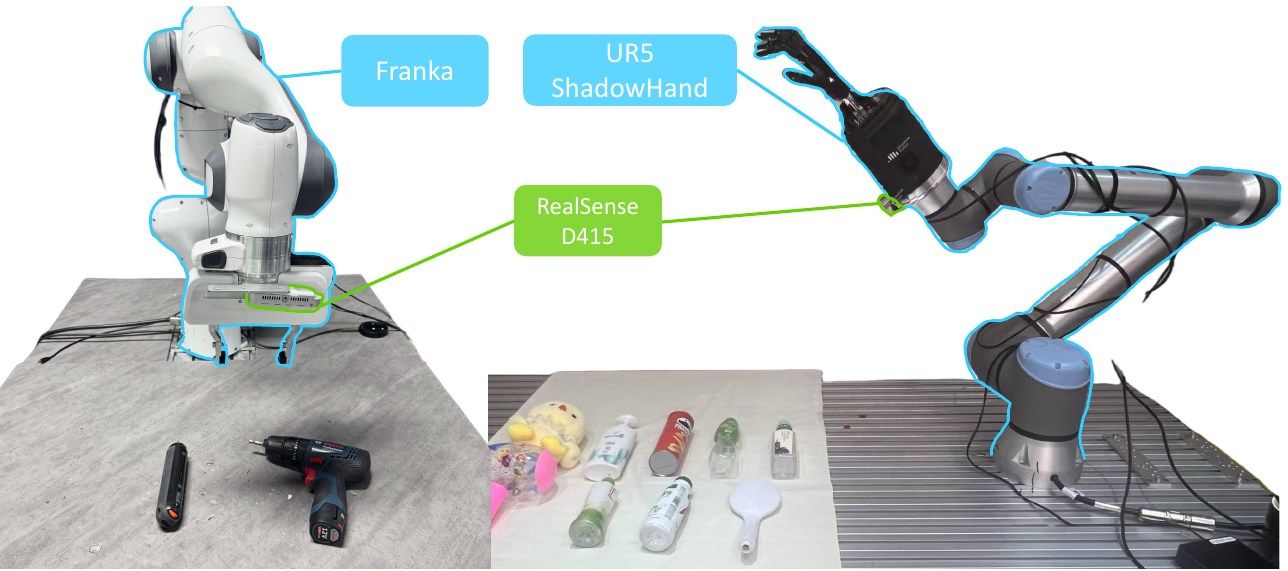}
\vspace{4pt}
\captionof{figure}{\textbf{Object Manipulation robot setup.}}
\label{fig:franka_setup}

\end{figure*}
\subsubsection{Real World Robot Setups}
We perform object manipulation tasks using the Franka Panda equipped with a gripper and the UR robot arm with a Shadow Hand. All the robot arms mount a RealSense D415 camera at their end for image capturing.
In \cref{fig:franka_setup}, we present the workspace and robotics for real-world object manipulation. We utilize only a single D415 camera. This setup significantly reduces the additional overhead associated with environmental setup and multi-camera calibration, and it is more readily reproducible.

\subsubsection{Language-Conditioned Grasping in Real-World Scenarios}
We evaluate robotic grasping tasks to test whether UniPart produces usable part masks on real sensor point clouds. We integrate UniPart with off-the-shelf grasp generators, GraspNet~\cite{Graspness24} and DexGraspNet2.0~\cite{zhang2024dexgraspnet}. Given a language query, UniPart predicts a target part point cloud, which is then used as the input to the grasp generator to execute a grasp.

\vspace{3pt}
\noindent\textbf{Quantitative Results}
We evaluate 20 real objects with language prompts that specify a target part. We report part segmentation accuracy by inspection of the predicted part, and we report grasp success as the fraction of successful grasps that land on the intended part. UniPart achieves 90.5\% part segmentation accuracy and the downstream pipeline achieves 85.0\% grasp success.

\vspace{3pt}
\noindent\textbf{Qualitative Results}
UniPart delineates parts such as handles and grips under sensor noise and occlusion, which supports its use as a perception module for language-conditioned grasping.

\begin{table}[t]
\centering
\caption{\textbf{Controlled study} evaluated on Objaverse-General (zero-shot mIoU). Both our dataset and architecture contribute significantly to the final performance.}
\label{tab:data_vs_model}
\small
\begin{adjustbox}{max width=\linewidth}
\begin{tabular}{llc}
\toprule
Model Architecture & Training Data & \texttt{mIoU} \\
\midrule
FIND3D & FIND3D Data & 30.75 \\
\method (Ours) & FIND3D Data & 35.27 \\
\midrule
FIND3D & LangPart-1M (Ours) & 38.42 \\
\method (Ours) & LangPart-1M (Ours) & \textbf{47.25} \\
\bottomrule
\end{tabular}
\end{adjustbox}

\end{table}

\subsection{Ablation Study}

\subsubsection{Controlled Study}
To strictly isolate the performance gains brought by our proposed architecture and our large-scale dataset, we conduct a controlled cross-evaluation against the FIND3D~\cite{ma2025find} baseline on the Objaverse-General zero-shot benchmark. We evaluate four settings: (1) FIND3D model trained on FIND3D data, (2) UniPart trained on FIND3D data, (3) FIND3D model trained on LangPart-1M, and (4) UniPart trained on LangPart-1M. 

As shown in \cref{tab:data_vs_model}, when restricted to the same FIND3D training data, UniPart still outperforms the FIND3D model by $14.69\%$ mIoU, demonstrating the superior cross-modal alignment capability of our additive text-injection Transformer. Conversely, when the FIND3D model is trained on our LangPart-1M dataset, its performance improves by $24.94\%$ mIoU, validating the high quality and scalability of our automated data engine. Ultimately, UniPart trained on LangPart-1M achieves the best performance, proving that our architecture is uniquely suited to absorb the rich supervision from large-scale text-part data.

\subsubsection{Ablation study on Pretraining}
To isolate the contribution of the proposed cross-modal pretraining, we conduct ablations under both partial and complete point-cloud settings. Although the pretraining stage is performed only on single-view partial point clouds (as it relies on point-to-image patch correspondences), we evaluate its effect on models trained and tested with partial inputs as well as on models operating on complete geometry. As shown in \cref{tab:pretrain_ab}, pretraining on single-view partial point clouds still improves fine-tuning on complete point clouds by transferring CLIP semantics into the 3D encoder. Gains are typically larger for partial inputs but remain consistent on complete geometry.
\begin{table*}[t!]
  \begin{minipage}[t]{0.55\textwidth}
    
    \centering
    \captionof{table}{Ablation study of pretraining for UniPart.}
    \vspace{5pt}
    \label{tab:pretrain_ab}
    \small
\begin{adjustbox}{max width=\linewidth}
    \begin{tabular}{lc}
    \toprule
     & \texttt{mIoU} \\ 
    \midrule
    with pretrain + partial PC & 33.56  \\
    with pretrain + complete PC & 27.17 \\
    without pretrain + partial PC & 26.34 \\
    without pretrain + complete PC & 23.39\\
    \bottomrule
    \end{tabular}
    \end{adjustbox}
  \end{minipage}
  \hfill
  \begin{minipage}[t]{0.42\textwidth}
    \centering
    \captionof{table}{Ablation study of multi-modal fusion.}
    \vspace{5pt}
    \label{tab:fusion}
    \small
\begin{adjustbox}{max width=\linewidth}
    \begin{tabular}{lc}
    \toprule
    Fusion Method & \texttt{mIoU} \\ 
    \midrule
    Cross-attn & 29.22  \\
    Multiplication & 28.13 \\
    Concat & 30.31 \\
    Addition & \textbf{33.56} \\
    \bottomrule
    \end{tabular}
    \end{adjustbox}
  \end{minipage}
  
\end{table*}

\subsubsection{Cross-Modal Fusion Choices}\label{app:fusion}
We further conduct an ablation study on the multi-modal fusion methods in UniPart, testing commonly used feature fusion techniques such as cross-attention, multiplication, addition, and concatenation, as shown in \cref{tab:fusion}. The results indicate that simple addition achieves the best performance. This may be attributed to the fact that instructions in the semantic domain are typically composed of short phrases or sentences, and the text CLS token already encodes sufficiently high-level semantic information.

\section{Conclusion and Future Work}
\label{sec:conclusion}

We introduced \textbf{UniPart} for zero-shot, language-grounded 3D part segmentation. Our goal is to provide a simple semantic interface that maps a natural language phrase to a fine-grained 3D part mask, which is a useful primitive for embodied interaction. A key enabler is \textbf{LangPart-1M}, a large-scale resource with 8 million text to part alignments on 160K+ objects, constructed with a multi-view pipeline that improves 3D consistency through text guided merging. We further curate \textbf{LangPart-4K} with manual verification and establish \textbf{LangPart-1K} for evaluation. UniPart, a feed-forward cross-modal 3D Transformer with layer-wise additive text conditioning, achieves strong generalization across open-vocabulary benchmarks and produces usable part masks on real sensor point clouds for language-conditioned grasping. Future work will extend part understanding beyond isolated objects and strengthen closed-loop integration with robotic control.

\bibliographystyle{assets/plainnat}
\bibliography{main}
\end{document}